\pdfoutput=1

\documentclass{article}

\usepackage{arxiv}

\usepackage[utf8]{inputenc} 
\usepackage[T1]{fontenc}    
\usepackage{url}            
\usepackage{booktabs}       
\usepackage{amsfonts}       
\usepackage{nicefrac}       
\usepackage{microtype}      
\usepackage{lipsum}		
\usepackage{graphicx}
\usepackage{natbib}
\usepackage{doi}

\usepackage{framed,multirow}

\usepackage{amssymb}
\usepackage{latexsym}

\usepackage{url}
\usepackage{xcolor}
\usepackage{tabularx}
\usepackage{multirow}
\usepackage{graphicx}

\usepackage{array}    
\usepackage{booktabs} 
\usepackage{pifont}
\usepackage[title]{appendix}

\usepackage{tikz}
\usetikzlibrary{shapes.geometric, arrows}

\definecolor{newcolor}{rgb}{.8,.349,.1}

\title{Mitigating the Influence of Domain Shift in Skin Lesion Classification: A Benchmark Study of Unsupervised Domain Adaptation Methods on Dermoscopic Images} 

\date{} 					
  \author{Sireesha Chamarthi \thanks{\texttt{Both authors contributed equally}}\\
	Data Analysis and Intelligence\\
	German Aerospace Center (DLR)\\
	Jena, Germany \\
	\texttt{Sireesha.Chamarthi@dlr.de} \\
   \And
  Katharina Fogelberg\footnotemark[1]\\
	Digital Biomarkers for Oncology \\
	German Cancer Research Center (DKFZ)\\
	Heidelberg, Germany \\
	\texttt{katharina.fogelberg@dkfz-heidelberg.de} \\
	\AND
	Roman C. Maron \\
        Digital Biomarkers for Oncology \\
	German Cancer Research Center (DKFZ) \\
	\And
	Titus J. Brinker \\
        Digital Biomarkers for Oncology \\
	German Cancer Research Center (DKFZ) \\
	\And
	Julia Niebling \\
        Data Analysis and Intelligence\\
	German Aerospace Center (DLR) \\
}

\renewcommand{\headeright}{Chamarthi et al, 2023}
\renewcommand{\undertitle}{Preprint}
\renewcommand{\shorttitle}{Mitigating the Influence of Domain Shift}

\hypersetup{
pdftitle={Mitigating the Influence of Domain Shift in Skin Lesion Classification: A Benchmark Study of Unsupervised Domain Adaptation Methods on Dermoscopic Images},
pdfsubject={cs.CV},
pdfauthor={Sireesha Chamarthi,Katharina Fogelberg},
pdfkeywords={domain shift, skin lesion classification},
}

\begin{document}
\maketitle

\begin{abstract}
The potential of deep neural networks in skin lesion classification has already been demonstrated to be on-par if not superior to the dermatologists’ diagnosis. However, the performance of these models usually deteriorates when the test data differs significantly from the training data (i.e. domain shift). This concerning limitation for models intended to be used in real-world skin lesion classification tasks poses a risk to patients. For example, different image acquisition systems or previously unseen anatomical sites on the patient can suffice to cause such domain shifts. Mitigating the negative effect of such shifts is therefore crucial, but developing effective methods to address domain shift has proven to be challenging. In this study, we carry out an in-depth analysis of eight different unsupervised domain adaptation methods to analyze their effectiveness in improving generalization for dermoscopic datasets. To ensure robustness of our findings, we test each method on a total of ten distinct datasets, thereby covering a variety of possible domain shifts. In addition, we investigated which factors in the domain shifted datasets have an impact on the effectiveness of domain adaptation methods. Our findings show that all of the eight domain adaptation methods result in improved AUPRC for the majority of analyzed datasets. Altogether, these results indicate that unsupervised domain adaptations generally lead to performance improvements for the binary melanoma-nevus classification task regardless of the nature of the domain shift. However, small or heavily imbalanced datasets lead to a reduced conformity of the results due to the influence of these factors on the methods' performance. 
\end{abstract}


\newpage

\section{Introduction}
\label{sec1}

Deep Neural Networks (DNNs) transformed machine learning by significantly improving predictive accuracy, even in complex real-world applications like skin cancer classification. Usually, DNNs are trained on large datasets, so they learn the representations effectively. Apart from that, the training dataset (or source) and the test dataset (or target) for classification models are drawn from the same distribution. However, in skin cancer classification, as well as in other real-world scenarios, the source and target domains are generally different. Even a small-scale deviation from the distribution of the training domain can result in unreliable and deteriorated predictions on the target domain \cite{yosinski2014transferable, NEURIPS2019_8558cb40, ADDA_2017, AFN_2019}. This deviation between datasets is commonly referred to as domain shift. In dermatology, these domain shifts can be caused by a combination of different factors, such as changes in the settings of an image acquisition system, view angle, lighting conditions in the examination room or the way the dermatoscope is positioned, among others.

As such domain shifts lead to a drop in performance, there exist different approaches to handle that problem, e.g. data augmentation \cite{yao2022improving}, domain generalization \cite{Wang_2021} and domain adaptation (DA) \cite{Wang_2018, Guo_2021}. Domain generalization and DA are closely related. While domain generalization methods do not access any data from the target domain, domain adaptation methods may make use of data from the target domain by definition. Nevertheless, all these approaches can only reduce, but not remove the discrepancy between domains \cite{yosinski2014transferable}.

Domain adaptation is typically applied in cases where the domain feature spaces and tasks remain the same while only the distributions differ between source and target datasets (presence of a domain shift) \cite{quinonero2008dataset, DAN_2015, DANN_2015}. Mainly this is done by either moment-matching methods or by adversarial learning \cite{CDAN_2018}. The knowledge transfer from source to target works via finding domain-invariant representations, which are used to bridge the discrepancy between domains \cite{pan2010survey}. 

Unsupervised Domain Adaptation (UDA) methods are well studied and established on multiple benchmark datasets (usually natural images), like Office-10, Caltech-10, Office-31, MNIST and SVHN datasets \citep{Patel_2015,Wang_2018, zhang2021survey}, but their performance is not verified on new tasks \cite{MCC_2020}. Therefore it may be more difficult to choose a proper method for real-life applications. Apart from this, existing medical images are mostly unlabeled as it is generally difficult to obtain labeled data in the medical field. For a sufficient ground truth (labels) for dermoscopic images, a biopsy of the human lesion needs to be performed. Therefore, the overall process of obtaining and reliably labeling dermoscopic data is labor-intensive. That is why further task-specific fine-tuning of DNN is time-consuming and difficult. These limitations can be addressed by utilizing specifically domain adaptation methods which are unsupervised \citep{Guan_2022}.

To our knowledge, there is no previous research that applies UDA methods as a benchmark on dermoscopic skin cancer datasets. Most existing works on domain adaptation assume their datasets to be domain shifted without quantifying it. However, in more complex tasks such as dermoscopic image classification, where even medical experts struggle to differentiate melanomas and nevi in particular situations, it is crucial to ensure that the datasets are truly domain shifted. In our previous work \cite{self2023domain}, we grouped and quantified domain shifted datasets for dermoscopic skin cancer classification, which we will use in this study. Additionally, other studies acknowledge their performance improvements without focusing on influential factors. However, we aim to identify possible factors for this performance improvement. Furthermore, other studies typically focus only on their benchmark and do not compare their results to other tasks, which can limit the generalizability of their findings. Therefore, while good performance on one method with one dataset or task may indicate its effectiveness, it does not guarantee the same performance improvement with other datasets or tasks.\\

Our contributions are the following:
\begin{itemize}
    \item we provide a comparative analysis of 8 UDA methods on 11 dermoscopic datasets with quantified (not assumed) biological and technical domain shifts.
    \item we identify dataset-specific factors on the performance of UDA methods.
    \item we compare our results to other benchmark domain adaptation datasets (e.g. Office-31).
\end{itemize}

This work is structured as follows:
First we discuss related works which focuses on UDA methods and benchmarking in \autoref{sec:Relevant}. In \autoref{sec:materialsmethods} we describe the used dermoscopic datasets and the UDA methods we compared in our analyses. We further explain our experimental settings. Finally, in \autoref{sec:results} we explain our results regarding different aspects of comparison. We discuss the influence of class imbalance, target dataset size, as well as the performance itself using the Area Under the Receiver Operating Characteristic (AUROC) and the Area Under the Precision-Recall Curve (AUPRC). We conclude the paper in \autoref{sec:conclusion} with our findings and discuss possible future research directions.

\section{Related work}
\label{sec:Relevant}

Ben-David et al., \cite{Ben_David_2009} pioneered domain adaptation theory and further classified DA methods to supervised and unsupervised approaches based on label availability. In Supervised Domain Adaptation (SDA) the model is trained on the source domain and tested on the target domain, both with labeled data. The most common approach for SDA is pretraining on the source domain and fine-tuning on the target domain. However, for the translation of medical applications into the clinic this approach is impractical and time-consuming, because it needs to be retrained for every new clinical scenario. The main goal of UDA is to enable the adaptation to new domains for better generalization by matching the marginal \cite{NIPS2006_a2186aa7, NIPS2007_be83ab3e, pan2010domain, gong2013connecting} or the conditional distributions \cite{pmlr-v28-zhang13d, courty2017joint} of the labeled source and unlabeled target domains. As dearth of labeled data is the most prominent issue in the medical field, UDA methods gained a lot of attention especially in medical image analysis \cite{Guan_2022}. Owing to advantages of UDA- over SDA methods, most of the existing DA research is focused on UDA. To enable adaptation from source domain to target domain, UDA methods have to meet two important criteria, namely transferability and discriminability \cite{BSP_2019}. The transferability of feature representations from source to target is the primary indicator for the performance of the model. Apart from this, the other key indicator is the ability to discriminate between the classes present in the domains. 
There are mainly two strategies to align feature distributions across domains: Moment matching and adversarial training \cite{MCC_2020}. 

\textbf{Moment matching methods} aim to decrease the distribution discrepancy  between the source and the target domain. This is achieved by matching the first/second order moments (as mean and covariance) of the activation distributions that are unique to each domain in the hidden activation space \cite{CMD_2017}. Multiple UDA methods have been developed based on moment matching, including Deep Adaptation Networks (\textit{DAN}) \cite{DAN_2015} which utilizes Maximum Mean Discrepancy (\textit{MMD}). An extension of \textit{DAN}, called Joint Adaptation Networks (\textit{JAN}) \cite{JAN_2017} has also been established. Apart from that, Correlation Alignment (\textit{Deep-CORAL}) \cite{CORAL_2016} is based on second order statistics of both the distributions. Another approach is \textit{CMD} \cite{CMD_2017}, which defines a new distance function based on probability distributions by moment sequences. Methods based on divergence are typically not very complicated, easier to train and do not require a lot of hyperparameter tuning for optimization. Additionally, they are computation efficient and are not in necessity of large datasets \cite{Zhao_2022}. However, the disadvantage of these types of methods is that they cannot be reliably used to achieve good performances on large datasets with more complex and diverse images. Also, they cannot be applied to other computer vision tasks, as semantic segmentation, because they do learn image-level, not pixel-level representations.

\textbf{Adversarial training methods} for domain adaptation learn domain-invariant features. For this, a domain discriminator is trained to differentiate between the source and the target domain by minimizing the classification error. At the same time, the feature representations learned by the classifier try to confuse the discriminator. One of the well-studied and most used adversarial methods is the Domain Adversarial Neural Network (\textit{DANN}) \cite{DANN_2015}. Apart from \textit{DANN}, there are other adversarial methods like Adversarial Discriminative Domain Adaptation (\textit{ADDA}) \cite{ADDA_2017} and  Maximum Classifier Discrepancy (\textit{MCD}) \cite{MCD_2018} which are developed as an extension to the \textit{DANN} approach. Typically, adversarial methods achieve better adaptations than moment matching methods and are the more dominant method \cite{AFN_2019}. They are very good in enhancing transferability of representations. Additionally, they have good computational efficiency and work across different kinds of datasets \cite{Zhao_2022}. Discriminative approaches are able to adapt well on larger domain shifts \cite{ADDA_2017}.
The disadvantage is that it can be difficult to optimize them. Also, in some cases they may perform poorly on small datasets, because these methods rely on the convergence of a min-max game. When having multimodal feature distributions it can be challenging for adversarial methods to adapt feature representations only \cite{NIPS2014_5ca3e9b1, arjovsky2017towards, CDAN_2018}. The discriminability of the learned representation happens only by minimizing the classification error on the source domain \cite{BSP_2019}. It cannot be guaranteed that the distributions are identical, even if the confusion of the discriminator was fully achieved \cite{pmlr-v70-arora17a, CDAN_2018}. 

There exist also \textbf{extensions to adversarial methods}, e.g. Batch Spectral Penalization ({\textit{BSP}) \cite{BSP_2019}, which can be used standalone or as a regularizer to another domain adaptation method. Also Minimum Class Confusion (\textit{MCC}) \cite{MCC_2020} can be used as a standalone adaptation method or additionally as a regularizer. The advantage of \textit{MCC} over \textit{BSP} is, that \textit{MCC} as a regularizer is not limited to adversarial methods.

Besides, there is also extensive research in the direction of \textbf{adversarial generative methods} which are based on GANs. They include a generator to create virtual images, while a discriminator tries to differentiate between real and generated images \cite{NIPS2014_5ca3e9b1}. The research in the area of conditional GANs \cite{condGan_2014} led to the development of methods like Conditional Adversarial Domain Adaptation ({\textit{CDAN}) \cite{CDAN_2018}. Although adversarial generative adaptation methods usually achieve good performances, they require largely scaled data for the generator to be trained properly. Furthermore, these methods need more computational resources, as well as hyperparameter tuning, which makes the optimization process more complex \cite{Zhao_2022}. Additionally, GANs show attractive visualizations, but they can be limited to small shifts \cite{ADDA_2017}. 

Due to the growing demand in adapting neural networks to unseen domains, there are other popular methods like Unsupervised Image-to-Image Translation Networks (\textit{UNIT}) \cite{UNIT_2017}, Generate to Adapt (\textit{GTA}) \cite{GTA_2018}, Cycle-Consistent Adversarial Domain Adaptation (\textit{CyCADA}) \cite{hoffman2018cycada} and Adaptive Feature Norm (\textit{AFN}) \cite{AFN_2019}.

It is important to note that this area of research is rapidly growing and new domain adaptation methods are emerging in a variety of fields, ranging from computer vision, natural language processing, video analysis to robotics. Their use-case is also not just limited to image classification tasks, but is extended to semantic segmentation, face recognition, object identification, image-to-image translation, person re-identification and image captioning among others \cite{Wang_2018}. Domain adaptation is also commonly used in medical image analysis. The leading application area of visual domain adaptation in medicine are brain images \cite{Guan_2022}, while there is also research on lungs, hearts, breasts, eyes and abdomen. Mostly, these applications use histological or microscopical images. 

We have noticed that there is limited work applying DA to dermoscopic images. Gu et al., \cite{Gu_2020} developed a two-step progressive adaptation method for task specific skin cancer classification. In their approach, they first trained a CNN on ImageNet and further fine-tuned it on an intermediate skin cancer dataset, before fine-tuning it again on another skin cancer dataset. Apart from that, Ahn et al., \cite{Ahn_2020} used a similar approach of training the model initially on ImageNet and fine-tuning it on medical images. They used context-based feature augmentation which uses additional information about the images. They experimented with medical image modality classification, a tuberculosis dataset, as well as with skin cancer datasets.

\begin{table*}[h!]
\centering
\begin{tabular}{lccccccc@{}}

\toprule[0.9pt]

Abbreviation & \begin{tabular}[c]{@{}c@{}}Origin \end{tabular} & \begin{tabular}[c]{@{}c@{}}Biological factors \end{tabular} &\begin{tabular}[c]{@{}c@{}}Melanoma\\  amount\end{tabular} & \begin{tabular}[c]{@{}c@{}}
Nevus\\  amount\end{tabular} & \begin{tabular}[c]{@{}c@{}}Total target \\ size\end{tabular} \\

\toprule[0.9pt]
H     & HAM & Age \textgreater 30, Loc. = Body (default)  & 465 (10\%)                     & 4234 (90\%)                 & 4699                  \\

HA & HAM & Age $\leq30$, Loc. = Body   & 25 (4\%)                     & 532 (96\%)                 & 557                  \\
HLH   & HAM & Age \textgreater 30, Loc. = Head/Neck                    & 99 (45\%)                     & 121 (55\%)                 & 220                  \\
HLP   & HAM & Age \textgreater 30, Loc. = Palms/Soles                    & 15 (7\%)                      & 203 (93\%)                 & 218  
    \\
B & BCN & Age \textgreater 30, Loc. = Body (default)                      &1918 (41\%)                      & 2721 (59\%)                & 4639                 \\
BA & BCN & Age $\leq30$, Loc. = Body                     & 71 (8\%)                     & 808 (92\%)                 & 879                  \\
BLH & BCN & Age \textgreater 30, Loc. = Head/Neck                     & 612 (66\%)                    & 320 (34\%)                 & 932     
    \\
BLP  & BCN & Age \textgreater 30, Loc. = Palms/Soles                    & 192 (65\%)                    & 105 (35\%)                 & 297                  \\
M  & MSK & Age \textgreater 30, Loc. = Body (default)                    & 565 (31\%)                    & 1282 (69\%)                & 1847                 \\
MA  & MSK & Age $\leq30$, Loc. = Body                    & 37 (8\%)                     & 427 (92\%)                 & 464                  \\
MLH & MSK & Age \textgreater 30, Loc. = Head/Neck                     & 175 (60\%)                    & 117 (40\%)                 & 292              
\\ 
\bottomrule[0.9pt]
\end{tabular}
\caption{Overview of the datasets used for our benchmark including dataset sizes and class distributions. H represents our source dataset. All following datasets are domain shifted datasets with respect to H.}
\label{table:divergencetable}
\end{table*}

UDA methods are typically compared against each other when a new method is proposed. In that comparison the works mostly focus on performance comparisons with respect to other state-of-the-art methods. Most of the UDA methods are evaluated on well-studied datasets like ImageNet, MNIST and Office-31, whereas their performance on other datasets is expected to change based on the available data and the domains present in them. Even these benchmark datasets are not analyzed for artefacts and duplicates present within the dataset. Ringwald et al. \cite{Ringwald_2021} analyzed frequently used UDA datasets and studied the systematic problems with regard to dataset setup and ambiguities. They established a clean Office-31 dataset for UDA algorithm comparisons. Hence, to verify the actual efficiency of the UDA methods, it is essential to study their performance on other, more real-world related datasets, as well. Peng et al., \cite{Peng_2018} introduced a benchmark dataset to evaluate the performance of UDA methods. They estimate the performance of the domain adaptation models to transfer knowledge from synthetic to real data. Also, Nagananda et al., \cite{Nagananda_2021} compared UDA methods on publicly available aerial datasets. In the medical field, Saat et al. \cite{Saat_2022} proposed a benchmark for UDA methods on brain Magnetic Resonance Imaging (MRI) - an image segmentation task. In their work, they compared UDA methods and evaluated the performance with respect to their baseline model. The source domain consists of MRI scans from multiple centers and different scanners. Whereas the target domain consists of MRI scans from a different dataset from a single center. We noticed that there is no extensive work on benchmarking UDA methods in particular for dermoscopic skin cancer image classification. 

\section{Materials and Methods}
\label{sec:materialsmethods}

\subsection{Datasets}
\label{ssec:datasets}

Even though some recent works used image datasets from skin lesions like MoleMap, HAM10k and ISIC \cite{Guan_2022} for their adaptation tasks, there is no study evaluating the actual and total domains present in these datasets or developing and evaluating public dermoscopic datasets particularly for domain adaptation techniques \citep{Saat_2022}. To overcome this limitation, we grouped and quantified potential technical and biological shifts in our previous work \citep{self2023domain} to obtain domain shifted dermoscopic datasets\footnote{\texttt{https://gitlab.com/dlr-dw/isic\char`_download}}. 
\autoref{table:divergencetable} provides a summary of the domains observed in the dermoscopic datasets.

As we are using unsupervised approaches, the source domain is labeled and these labels are used for the classification at the end. Hence, it is essential to have a dataset that can be divided into train and test without data leakage, which is not always straightforward for dermatology datasets due to duplicated lesion images. Apart from that, in domain adaptation works, the methods are evaluated from one domain to another (domain A to domain B) and are also tested in the opposite direction (domain B to domain A) \cite{DANN_2015}. However, recent works stated that the performance of UDA methods is negatively affected by poor data quality and duplicates in the datasets \cite{Ringwald_2021}. This can be a difficulty when using the publicly available ISIC archive images as they contain duplicates which are not necessarily marked as such \cite{cassidy2022analysis}. 

We chose a representative and large subset of HAM, dataset \textit{H} (\autoref{table:divergencetable}), as our only source domain for the adaptation process. For this, we used the lesion ID's present in HAM10k to remove the duplicates in the dataset \citep{tschandl2018ham10000}. Therefore, the adaptation was done in one direction only, using sub-datasets \textit{HA}, \textit{HLH}, \textit{HLP}, as well as BCN20k \citep{combalia2019bcn20000} and MSK \citep{cassidy2022analysis} datasets exclusively as target domains.

\subsection{UDA Methods}
\label{ssec:methods}

Overall, we focus on single-source, single-target, homogeneous adaptation without target labels. This means that there is one fully labeled source domain and one unlabeled target domain within the same modality, and that the source and target domains share the same classes. 

We selected eight state-of-the-art UDA methods (\autoref{table:UDA_methods}), which were selected based on different types of UDA methods, computational efficiency and a good performance on different established datasets.

\begin{table}[h!]
\centering
\begin{tabular}{lccccc@{}}

\toprule[0.9pt]

UDA method & \begin{tabular}[c]{@{}c@{}}Type\end{tabular}  \\

\toprule[0.9pt]

DAN \cite{DANN_2015}     & Moment matching              \\
JAN \cite{JAN_2017}    & Moment matching            \\
DANN \cite{DANN_2015}   & Adversarial training            \\
ADDA \cite{ADDA_2017}   & Adversarial training             \\
BSP \cite{BSP_2019}   & Extension of adversarial              \\
MCC \cite{MCC_2020}     & Extension of adversarial                \\
CDAN \cite{CDAN_2018}    & Adversarial generative                \\
AFN \cite{AFN_2019}    & Other              \\
\bottomrule[0.9pt]
\end{tabular}
\caption{Our selection of eight state-of-the-art UDA methods for the benchmark study. The methods are of different types of strategies.}
\label{table:UDA_methods}
\end{table}

\subsection{Experimental setup}
\label{ssec:experimental}

\begin{table*}[h!]
\centering
\begin{tabular}{|c|cccccccccccc@{}}
\cmidrule[0.9pt](l){3-12}
\multicolumn{1}{c}{} & \multicolumn{7}{r@{}}{Domain shifted dataset}\\
\cmidrule(l){3-12}
\multicolumn{2}{c}{}& HA & HLH & HLP & B & BA & BLH & BLP & M & MA & MLH  \\
\cmidrule(l){2-12}
\multicolumn{2}{c}{}Mel (\%) & 4 & 45 & 7 & 41 & 8 & 66 & 65 & 31 & 8 & 60  \\
\cmidrule(l){2-12}
\multirow{5}{*}{\rotatebox[origin=c]{90}{(UDA) method\hspace{1.2cm}}}
& Src & 0.14{\tiny$\pm$0.02} & 0.69{\tiny$\pm$0.04} & 0.37{\tiny$\pm$0.15} & 0.57{\tiny$\pm$0.02} & 0.19{\tiny$\pm$0.06} & 0.73{\tiny$\pm$0.03} & 0.77{\tiny$\pm$0.05} & 0.34{\tiny$\pm$0.01} &  0.15{\tiny$\pm$0.04} & 0.68{\tiny$\pm$0.03}\\ [1.2ex]
& DAN &  0.12{\tiny$\pm$0.02} & 0.77{\tiny$\pm$0.02} & 0.47{\tiny$\pm$0.14} & 0.60{\tiny$\pm$0.04} & 0.20{\tiny$\pm$0.02} & 0.78{\tiny$\pm$0.01} & 0.83{\tiny$\pm$0.03} & 0.37{\tiny$\pm$0.03} &  0.13{\tiny$\pm$0.03} & 0.69{\tiny$\pm$0.03}  \\ [0.2ex]
& JAN & 0.15{\tiny$\pm$0.04} & 0.82{\tiny$\pm$0.05} & 0.56{\tiny$\pm$0.08} & 0.72{\tiny$\pm$0.02} & 0.34{\tiny$\pm$0.02} & 0.85{\tiny$\pm$0.02} & 0.82{\tiny$\pm$0.03} & 0.44{\tiny$\pm$0.03} &  0.14{\tiny$\pm$0.01} & \textbf{0.73{\tiny$\pm$0.03}}  \\ [0.2ex]
& DANN &  0.17{\tiny$\pm$0.01} & 0.81{\tiny$\pm$0.04} & 0.55{\tiny$\pm$0.07} & 0.74{\tiny$\pm$0.02} & 0.32{\tiny$\pm$0.03} & 0.85{\tiny$\pm$0.01} & 0.84{\tiny$\pm$0.01} & 0.44{\tiny$\pm$0.01} &  \textbf{0.18{\tiny$\pm$0.04}} & 0.72{\tiny$\pm$0.02} \\ [0.2ex]
& ADDA & \textbf{0.18{\tiny$\pm$0.06}} & 0.81{\tiny$\pm$0.02} & 0.55{\tiny$\pm$0.03} & 0.74{\tiny$\pm$0.01} & \textbf{0.36{\tiny$\pm$0.03}} & \textbf{0.87{\tiny$\pm$0.01}} & 0.83{\tiny$\pm$0.02} & 0.44{\tiny$\pm$0.02} &  0.17{\tiny$\pm$0.03} & \textbf{0.73{\tiny$\pm$0.03}} \\ [0.2ex]
& CDAN & 0.14{\tiny$\pm$0.02} & 0.82{\tiny$\pm$0.03} & 0.54{\tiny$\pm$0.06} & 0.73{\tiny$\pm$0.02} & 0.33{\tiny$\pm$0.02} & 0.85{\tiny$\pm$0.01} & 0.84{\tiny$\pm$0.02} & \textbf{0.47{\tiny$\pm$0.02}} &  0.14{\tiny$\pm$0.01} & \textbf{0.73{\tiny$\pm$0.02}}  \\ [0.2ex]
& BSP & 0.16{\tiny$\pm$0.03} & 0.82{\tiny$\pm$0.02} & \textbf{0.65{\tiny$\pm$0.04}} & \textbf{0.75{\tiny$\pm$0.01}} & 0.34{\tiny$\pm$0.05} & 0.86{\tiny$\pm$0.01} & 0.83{\tiny$\pm$0.02} & 0.46{\tiny$\pm$0.02} &  0.17{\tiny$\pm$0.03} & \textbf{0.73{\tiny$\pm$0.01}} \\ [0.2ex]
& AFN & 0.11{\tiny$\pm$0.02} & \textbf{0.83{\tiny$\pm$0.02}} & 0.57{\tiny$\pm$0.13} & 0.73{\tiny$\pm$0.01} & 0.35{\tiny$\pm$0.03} & 0.84{\tiny$\pm$0.01} & \textbf{0.86{\tiny$\pm$0.02}} & 0.43{\tiny$\pm$0.02} &  0.16{\tiny$\pm$0.02} & 0.71{\tiny$\pm$0.02}  \\ [0.2ex]
& MCC & 0.15{\tiny$\pm$0.04} & \textbf{0.83{\tiny$\pm$0.07}} & 0.57{\tiny$\pm$0.05} & 0.69{\tiny$\pm$0.02} & 0.30{\tiny$\pm$0.08} & 0.83{\tiny$\pm$0.01} & 0.81{\tiny$\pm$0.04} & 0.41{\tiny$\pm$0.02} &  0.14{\tiny$\pm$0.03} & 0.71{\tiny$\pm$0.03}   \\ [0.2ex]
\cmidrule[0.9pt](l){2-12}
\end{tabular}
\caption{Comparison of AUPRC results across different datasets and UDA methods. The columns represent the domain shifted target datasets for the source dataset \textit{H} (not listed here). Each row represents the results for a particular UDA method, with the first row indicating the results for the unadapted baseline method (\textit{Src}). The best-performing UDA method for each dataset is highlighted in bold. The percentage for each dataset shows the ratio of melanoma in that dataset, which serves as the baseline for AUPRC. The source dataset \textit{H} contains only 10\% melanoma.}
\label{table:prauc}
\end{table*}

It is difficult to decide which UDA method is generally better compared to others in terms of design or performance. The key characteristic that determines the strength of a UDA method is its ability to transfer feature representations from a source- to a target domain. For this reason, we compare all UDA results to our unadapted baseline method (\textit{Src}) trained on source dataset \textit{H}, which is a basic ResNet50 model \cite{He_2016} pretrained on ImageNet. The other performance characteristic is the discriminability between the classes within domains. We evaluate how well the model is able to discriminate between melanoma and nevus in our binary classification task. For this we follow standard evaluation protocols for unsupervised domain adaptation \cite{DANN_2015, JAN_2017}. For all experiments we used an initial learning rate (LR) of 0.01 with a weight decay of 1e-3 and a LR-decay 0.75. The used momentum was 0.9 and gamma was 0.001. We set the epochs to 20 and the batch size to 16. The comparison is based on an existing repository\footnote{\texttt{https://github.com/thuml/Transfer-Learning-Library}} which already implemented a variety of methods. It is open-source and has been established on multiple popular datasets, e.g. MNIST, Office-31 and DomainNet \cite{jiang2022transferability,tllib}. We modified the library for our classification on dermoscopic images. 

In a typical dermoscopic dataset, the presence of melanoma, in comparison to nevus images, is very low, as can be seen in \autoref{table:divergencetable}. In our analysis, we consider melanoma as the positive and nevus as the negative class. When it comes to a clinical translation of a skin lesion diagnostic system, both, True Positives and True Negatives are considered very important. Therefore, we focused on AUROC and AUPRC as evaluation metrics. The advantage of these two metrics is that they are both threshold-free. That means, that they can provide an overview of the performance range with different dataset-splits into positively and negatively predicted classes \cite{saito2015precision}. Also Zhang et al. \cite{Zhang_ADADiag_2022} used AUROC and AUPRC for the evaluation of their domain adaptation results in a recent work. 

AUROC as a standalone metric can be misleading in imbalanced tasks, because the score can be better than random guessing (baseline=0.5), but still misclassify the minority class. On the contrary, AUPRC is tailored for such imbalanced cases, but may mislead in balanced cases or where the negatives are rare. When using only AUPRC, it can be difficult to compare results across datasets with different class ratios and dataset sizes. The reason for this is the varying baseline of this metric, as it is dependent on the ratio of the positive class. Therefore, we computed both metrics, while also focusing on the AUPRC improvement (in \%) compared to the unadapted baseline method (\textit{Src}), as suggested by Zhang et al. \citep{Zhang_2022}. With this approach, the results can be compared across methods and datasets equally.

For the experiments, we included a weighted random sampler to maintain equal class ratios per batch during model training. We also adopted five-fold cross validation to use all images of the available datasets. From each fold we selected the best epoch (out of 20) and averaged the results. Additionally, we ran the experiments with five seeds to observe the variability of the results over different runs. For the end results we averaged the values over five seeds. The seeding makes the performance results more robust and that way shows more realistic values.

\section{Results and Discussion}
\label{sec:results}

\begin{figure*}[htb]
\begin{minipage}[b]{0.95\textwidth}
  \centering
  \centerline{\includegraphics[width=\textwidth]{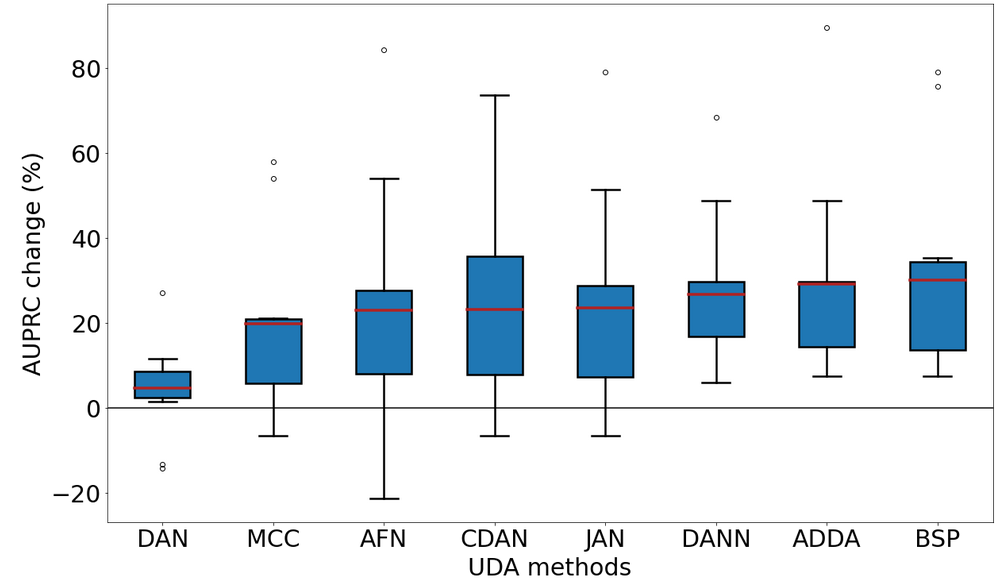}}
  \medskip
\end{minipage}
\caption{Comparison of UDA methods with respect to AUPRC change. The red line represents the mean (of the performance on all datasets) and the black dots are outliers. The black line shows the baseline at 0\% performance improvement. The performance improvement was calculated over five seeds and averaged over 10 datasets. The UDA methods are ordered in the increasing order of the mean AUPRC change on the x-axis. The numbers in the brackets (x-axis) represent for how many datasets out of 10 this particular method improved the performance.}
\label{fig:mean_impr_udamethods}
\end{figure*}

If a UDA method performs well on one dataset, it does not guarantee similar performance on other datasets. As discussed in \autoref{sec:materialsmethods}, we have selected 8 state-of-the-art UDA methods to evaluate the performance on the domains present in our 11 dermoscopic datasets. We compared all adaptation methods with our non-adapted baseline approach.

\subsection{Benchmarking UDA methods on dermoscopic datasets}
\label{ssec:analysisperformance}


As stated in \autoref{ssec:experimental}, we computed AUROC and AUPRC scores for different datasets and methods, which can be seen in \autoref{table:auroc} and \autoref{table:prauc}, respectively. To provide a better understanding of our comparison and to demonstrate the quantified changes compared to \textit{Src} method, we also looked at the AUROC- (\autoref{table:auroc_imp}) and AUPRC (\autoref{table:prauc_imp}) improvement (in \%). In these tables, negative values, which indicate performance degradation after domain adaptation, occur rarely.

Our results indicate, that all selected UDA methods achieve a performance improvement (in \%) compared to \textit{Src} method over most available domain shifted datasets (\autoref{fig:mean_impr_udamethods}). \textit{BSP}, \textit{ADDA} and \textit{DANN}, which are all adversarial types of techniques, achieve the largest performance improvement. According to our results, these three UDA methods were able to improve the performance of 10 out of 10 domain shifted dermoscopic datasets. The moment matching method \textit{DAN} did not exhibit an improvement in performance overall.

The performance change (in \%) of each individual domain shifted dataset per method is represented in \autoref{fig:prauc_change}. For this overview, we combine performance change, melanoma ratio and target dataset size in one figure. While the performance is demonstrated in the upper point plot, the melanoma ratio per dataset can be observed in the lower illustration. In both sub-figures the domain shifted datasets on the x-axis are ordered by target dataset size in an ascending order from left to right. The largest improvements are achieved on dataset \textit{BA} using either \textit{ADDA} or \textit{AFN} as the UDA method. However, although \textit{BA} has the highest improvement, it has also a high variance between the methods' results, which ranges from 5.26\% to 89.47\%. All UDA methods, except for \textit{DAN}, achieved maximum performance improvement at least for one dataset, as shown in \autoref{table:prauc_imp}. It is also noteworthy that the \textit{MLH} dataset posed the greatest challenge for adaptation, as all UDA methods seem to encounter difficulties with it (\autoref{fig:prauc_change}).

\begin{figure*}[htb]
\centering
\begin{minipage}[b]{0.9\textwidth}
  \centering
  \centerline{\includegraphics[width=\textwidth]{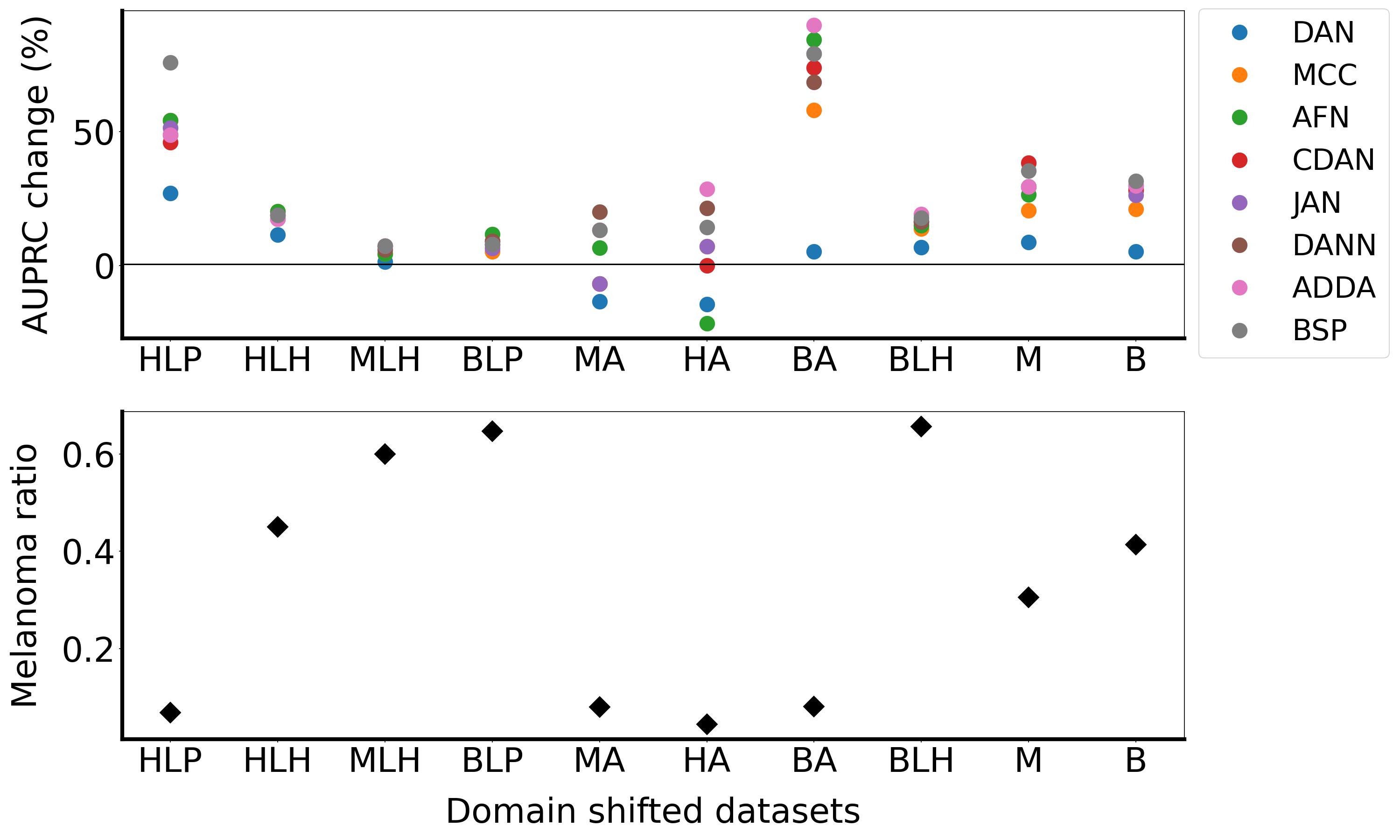}}
  \medskip
\end{minipage}
\caption{Change in AUPRC (in \%) with respect to the unadapted baseline model (\textit{Src}). Individual UDA methods (color-coded) are illustrated across all domain shifted datasets (x-axis). The upper panel of the figure shows the AUPRC change (in \%). The black line at 0 on the y-axis highlights the methods which show a performance degradation or no improvement (w.r.t the unadapted baseline method). The lower panel shows the melanoma ratio for each dataset. The datasets on the x-axis are ordered by total target size in an ascending manner.}
\label{fig:prauc_change}
\end{figure*}

\subsection{Influential factors on the performance improvement}
\label{ssec:factors}

Our analysis revealed that the amount of melanoma images in the target datasets affects the performance of UDA methods, as demonstrated in \autoref{table:prauc} and \autoref{fig:prauc_change}. Several datasets, including \textit{HA}, \textit{HLP}, \textit{MA}, and \textit{BA}, have a low number of melanoma cases and also represent larger disparities between the results of all UDA methods (\autoref{table:prauc_imp}). Datasets \textit{HLH}, \textit{B} and \textit{M} have a more balanced distribution between both classes (\autoref{table:divergencetable}). Adversarial methods and their extensions appear to perform better for such imbalanced datasets. For instance, \textit{ADDA} is the most effective UDA method for dataset \textit{HA}, which has the lowest melanoma sample size of 4\%. On the other hand, some datasets such as \textit{BLP}, \textit{BLH}, and \textit{MLH} are dominated by melanoma cases and therefore all methods show similar improvement in AUPRC scores. Although no improvement can be detected from the unadapted baseline, we observe that most methods agree with each other when it comes to datasets with a high melanoma ratio (\autoref{fig:prauc_change}). It is worth noting that \textit{MA}, \textit{HA}, and \textit{BA} datasets contain images of skin lesions from patients below the age of 30. These datasets include both, young patients and children, as we have previously noted in our work \cite{self2023domain}. Diagnosing melanoma in children is a unique challenge in clinical diagnosis, as they do not show typical ABCDE features \citep{duarte2021clinical} used to identify melanomas in adults due to their different appearance \citep{scope2016study}. This may result in a lower performance improvement after adaptation.

To achieve performance improvements in UDA methods, it is necessary to have a large dataset available for the training process of the adaptation method \cite{fewshotUDA}. As shown in \autoref{table:prauc} and \autoref{fig:prauc_change}, for the larger datasets \textit{M} and \textit{B}, most of the methods (except for \textit{DAN}) showed higher improvement in performance compared to other datasets. Interestingly, these two datasets have a balanced class distribution. However, the overall performance is also linked to the melanoma ratio in the dataset. An exception to this observation is dataset \textit{HLP}  where most methods show agreement despite the small dataset size and low melanoma ratio. We assume this is because of the relative similarity of the target dataset to the source dataset (\textit{H}). In our previous analysis \cite{self2023domain}, we found that \textit{HLP} is one of the most similar datasets to \textit{H} in terms of melanoma images, as measured by cosine similarity and JS-divergence. It is also worth noting that for this dataset, the variation between the least performing \textit{DAN} and the best performing \textit{BSP} method is also high.

Regarding the AUROC and AUPRC metrics, it appears that adversarial methods and their extensions generally outperform moment matching methods, with the exception of \textit{JAN} for the \textit{MLH} dataset in terms of AUPRC score. This finding is consistent with existing literature indicating that adversarial methods tend to perform better than moment matching methods \cite{AFN_2019}.

In summary, there are various factors that contribute to the performance of UDA methods, such as melanoma ratio, target size, and how similar or dissimilar datasets are with respect to the source dataset \textit{H}. However, it remains unclear which factor or combination of factors has the greatest influence on performance improvement with UDA. Therefore, it is essential to continue investigating and exploring these factors to better understand their impact on UDA performance.

\subsection{Performance of UDA methods on non-dermoscopic datasets}

Another reason for the usage of AUROC for evaluation is that various works on UDA methods compare their results either with AUROC or accuracy. In particular in the medical field it is common practice to compare methods based on AUROC scores \cite{purushotham2017variational, Zhou_2022, Feng_2023}. Most of the domain adaptation studies use accuracy as their metric for comparisons of methods, but none of these studies discuss the possible imbalance in their datasets.

Typical datasets used for domain adaptation tasks are, for instance, Digits or Office-31. These images are easier to adapt to and differ a lot more than dermoscopic images do. Moreover, benchmark datasets for UDA are typically large, have almost balanced classes and the classification ability can even be validated by non-expert humans. On the contrary, dermoscopic images look very similar, making the task not only difficult for medical experts, but also for the neural network. Backgrounds can contain complex structures, which are artefacts used by the neural networks for training, such as black borders, hair or skin colour. 

When comparing all used methods, DAN performs poorly in our dermoscopic scenario, as well as in other adaptation tasks \cite{DAN_2015, DANN_2015, JAN_2017, ADDA_2017, CDAN_2018, BSP_2019, AFN_2019, MCC_2020}. Our selection of UDA methods are benchmarked on Office-31-, Office-Home, ImageCLEF-DA- and VisDA17-datasets. It is worth noting that not all UDA methods are compared in each work and on each of these datasets. Therefore, a fair comparison of non-dermoscopic results is not possible.

We can observe, that the adversarial UDA methods, namely \textit{BSP} and \textit{ADDA}, which are overall performing better in our dermoscopic scenario, also perform very well in other image classification tasks. According to the authors of \cite{BSP_2019}, their method \textit{BSP} specifically boosts the performance on relatively difficult tasks, where the source domain is quite small. \textit{ADDA} was not often compared in these works, but it is outperformed by \textit{CDAN} in Office-31 adaptation. For all other methods, it is not possible to provide a clear order of performance improvements as they are compared on different tasks and with different methods. \textit{DAN}, \textit{DANN} and \textit{JAN} are outperformed by all mentioned methods in Office-31-, ImageCLEF-DA-, Office-Home- and VisDA17-adaptation. \textit{CDAN} is outperformed in Office-Home-, Office-31- and VisDA17-adaptation by \textit{AFN}, \textit{MCC} and \textit{BSP+CDAN}.

\section{Conclusion}
\label{sec:conclusion}
This is the first work benchmarking UDA methods on dermoscopic datasets. In summary, our work enables the reproducibility of results and interpretation due to the utilization of publicly available dermoscopic datasets. Further, the domain shifts in between the datasets were quantified, unlike for most used benchmark datasets. Our analysis shows that all selected UDA methods from different technical approaches improve the performance for most dataasets compared to the unadapted technique, although in different extents. 

We have additionally conducted a comparative analysis on the dependency of the performance of UDA methods on dataset-dependent features like class imbalance and target dataset size. We have noticed that the overall performance of UDA methods depends on various combinations of factors.

 Moreover, we compared how our chosen UDA methods perform for other typical benchmark adaptation tasks in comparison to our dermoscopic datasets. In general, the UDA methods that performed well on dermoscopic datasets are also the best performing methods on other non-dermoscopic tasks. 

In our analysis our aim was to compare different UDA methods on the same (well-established) ResNet-50 backbone. However, the selection of the backbone has an impact on the overall performance \cite{Feng_2023}. One further extension can be the comparison of different backbones and how it influences the performances. Another important future aspect is to investigate the intensity of performance degradation when reducing the target dataset size. It is worth exploring the field of few/one/zero-shot domain adaptation. Another possible area of interest is multi-source domain adaptation (multiple modalities) on dermoscopic datasets, where e.g. dermoscopic and clinical lesion images can be included. UDA could be recommended for adversarial methods, as these have consistently shown the most improvements. Ideally, the datasets should be balanced, because a low melanoma ratio was indicative of a high performance variance, thus making the applied methods riskier.


%

%

\newpage
\section*{Acknowledgements}
This research is funded by the \textit{Helmholtz Artificial Intelligence Cooperation Unit} [grant number ZT-I-PF-5-066].

The Helmholtz AI funding enabled the close cooperation between DKFZ and DLR, which leads to an interdisciplinary exchange between two research groups and thereby enables the integration of novel perspectives and experiences.

\bibliographystyle{abbrvnat}
\bibliography{main}


\appendix

\onecolumn
\section{}
\vspace{-15pt}
\setcounter{figure}{0} 
\setcounter{table}{0} 

\begin{table*}[h!]
\centering
\begin{tabular}{|c|cccccccccccc@{}}
\cmidrule[0.9pt](l){3-12}
\multicolumn{1}{c}{} & \multicolumn{7}{r@{}}{Domain shifted dataset}\\
\cmidrule(l){3-12}
\multicolumn{2}{c}{}& HA & HLH & HLP & B & BA & BLH & BLP & M & MA & MLH\\
\cmidrule(l){2-12}
\multirow{5}{*}{\rotatebox[origin=c]{90}{(UDA) method\hspace{1.2cm}}}
& Src & 0.65{\tiny$\pm$0.04} & 0.74{\tiny$\pm$0.05} & 0.82{\tiny$\pm$0.08} & 0.65{\tiny$\pm$0.01} & 0.58{\tiny$\pm$0.05} & 0.61{\tiny$\pm$0.03} & 0.62{\tiny$\pm$0.07} & 0.51{\tiny$\pm$0.01} &  \textbf{0.60{\tiny$\pm$0.05}} & 0.57{\tiny$\pm$0.03}\\ [1.2ex]
& DAN &  0.64{\tiny$\pm$0.06} & 0.79{\tiny$\pm$0.02} & 0.85{\tiny$\pm$0.06} & 0.67{\tiny$\pm$0.02} & 0.59{\tiny$\pm$0.02} & 0.65{\tiny$\pm$0.02} & 0.70{\tiny$\pm$0.06} & 0.52{\tiny$\pm$0.02} &  0.52{\tiny$\pm$0.05} & 0.59{\tiny$\pm$0.04}  \\ [0.2ex]
& JAN & 0.71{\tiny$\pm$0.04} & 0.85{\tiny$\pm$0.03} & 0.92{\tiny$\pm$0.02} & 0.76{\tiny$\pm$0.01} & 0.69{\tiny$\pm$0.01} & 0.74{\tiny$\pm$0.02} & 0.69{\tiny$\pm$0.04} & 0.62{\tiny$\pm$0.03} &  0.55{\tiny$\pm$0.06} & 0.61{\tiny$\pm$0.02}  \\ [0.2ex]
& DANN &  0.73{\tiny$\pm$0.03} & 0.84{\tiny$\pm$0.03} & 0.91{\tiny$\pm$0.03} & \textbf{0.78{\tiny$\pm$0.01}} & 0.67{\tiny$\pm$0.02} & 0.75{\tiny$\pm$0.01} & 0.72{\tiny$\pm$0.02} & 0.62{\tiny$\pm$0.01} &  \textbf{0.60{\tiny$\pm$0.03}} & 0.62{\tiny$\pm$0.03} \\ [0.2ex]
& ADDA & \textbf{0.74{\tiny$\pm$0.06}} & 0.84{\tiny$\pm$0.01} & 0.92{\tiny$\pm$0.01} & \textbf{0.78{\tiny$\pm$0.01}} & 0.68{\tiny$\pm$0.04} & \textbf{0.77{\tiny$\pm$0.01}} & 0.70{\tiny$\pm$0.03} & 0.62{\tiny$\pm$0.01} &  0.59{\tiny$\pm$0.03} & \textbf{0.63{\tiny$\pm$0.03}} \\ [0.2ex]
& CDAN & 0.71{\tiny$\pm$0.05} & 0.85{\tiny$\pm$0.01} & 0.90{\tiny$\pm$0.03} & 0.77{\tiny$\pm$0.01} & 0.68{\tiny$\pm$0.03} & 0.75{\tiny$\pm$0.02} & 0.71{\tiny$\pm$0.02} & \textbf{0.64{\tiny$\pm$0.02}} &  0.56{\tiny$\pm$0.02} & 0.62{\tiny$\pm$0.03}  \\ [0.2ex]
& BSP & 0.72{\tiny$\pm$0.03} & 0.84{\tiny$\pm$0.01} & \textbf{0.94{\tiny$\pm$0.02}} & \textbf{0.78{\tiny$\pm$0.01}} & \textbf{0.70{\tiny$\pm$0.04}} & 0.75{\tiny$\pm$0.02} & 0.71{\tiny$\pm$0.02} & \textbf{0.64{\tiny$\pm$0.02}} &  0.56{\tiny$\pm$0.02} & 0.62{\tiny$\pm$0.02} \\ [0.2ex]
& AFN & 0.67{\tiny$\pm$0.03} & 0.85{\tiny$\pm$0.02} & 0.92{\tiny$\pm$0.05} & 0.76{\tiny$\pm$0.01} & 0.66{\tiny$\pm$0.01} & 0.73{\tiny$\pm$0.01} & \textbf{0.73{\tiny$\pm$0.03}} & 0.60{\tiny$\pm$0.01} &  0.55{\tiny$\pm$0.02} & 0.60{\tiny$\pm$0.02}  \\ [0.2ex]
& MCC & 0.70{\tiny$\pm$0.01} &\textbf{0.87{\tiny$\pm$0.03}} & \textbf{0.94{\tiny$\pm$0.02}} & 0.76{\tiny$\pm$0.01} & 0.68{\tiny$\pm$0.03} & 0.73{\tiny$\pm$0.01} & 0.70{\tiny$\pm$0.04} & 0.61{\tiny$\pm$0.03} &  0.58{\tiny$\pm$0.08} & 0.59{\tiny$\pm$0.02}   \\ [0.2ex]
\cmidrule[0.9pt](l){2-12}
\end{tabular}
\caption{Comparison of AUROC results across different datasets and UDA methods. The columns represent the domain shifted target datasets for the source dataset \textit{H} (not listed here). Each row represents the results for a particular UDA method, with the first row indicating the results for the unadapted baseline method (\textit{Src}). The best-performing UDA method for each dataset is highlighted in bold.}
\label{table:auroc}
\end{table*}

\begin{table*}[h!]
\centering
\begin{tabular}{|c|cccccccccccc@{}}
\cmidrule[0.9pt](l){3-12}
\multicolumn{1}{c}{} & \multicolumn{7}{r@{}}{Domain shifted dataset}\\
\cmidrule(l){3-12}
\multicolumn{2}{c}{}& HA & HLH & HLP & B & BA & BLH & BLP & M & MA & MLH\\
\cmidrule(l){2-12}
\multirow{5}{*}{\rotatebox[origin=c]{90}{UDA method\hspace{0.8cm}}}
& DAN &  -1.54 & 6.76 & 3.66 & 3.08 & 1.72 & 6.56 & 12.9 & 1.96 & -13.33 & 3.51 \\ [0.2ex]
& JAN &  9.23 & 14.86 & 12.20 & 16.92 & 18.97 & 21.31 & 11.29 & 21.57 &  -8.33 & 7.02 \\ [0.2ex]
& DANN &  12.31 & 13.51 & 10.98 & \textbf{20.00} & 15.52 & 22.95 & 16.13 & 21.57 &  \textbf{0} & 8.77 \\ [0.2ex]
& ADDA & \textbf{ 13.85} & 13.51 & 12.20 & \textbf{20.00} & 17.24 & \textbf{26.23} & 12.90 & 21.57 & -1.67 & \textbf{10.53} \\ [0.2ex]
& CDAN &  9.23 & 14.86 & 9.76 & 18.46 & 17.24 & 22.95 & 14.52 & \textbf{25.49} & -6.67 & 8.77 \\ [0.2ex]
& BSP & 10.77 & 13.51 & \textbf{14.63} & \textbf{20.00} & \textbf{20.69} & 22.95 & 14.52 & \textbf{25.49} & -6.67 & 8.77 \\ [0.2ex]
& AFN & 3.08 & 14.86 & 12.20 & 16.92 & 13.79 & 19.67 & \textbf{17.74} & 17.65 &  -8.33 & 5.26 \\ [0.2ex]
& MCC & 7.69 & \textbf{17.57} & \textbf{14.63} & 16.92 & 17.24 & 19.67 & 12.9 & 19.61 &  -3.33 & 3.51 \\ [0.2ex]
\cmidrule[0.9pt](l){2-12}
\end{tabular}
\caption{Comparison of change in AUROC results for each UDA method with respect to the unadapted baseline method (\textit{Src}) for different datasets. H is the source dataset and the target datasets are listed in columns. The rows represent the improvements (in \%) for each UDA-method.}
\label{table:auroc_imp}
\end{table*}

\begin{table*}[h!]
\centering
\begin{tabular}{|c|cccccccccccc@{}}
\cmidrule[0.9pt](l){3-12}
\multicolumn{1}{c}{} & \multicolumn{7}{r@{}}{Domain shifted dataset}\\
\cmidrule(l){3-12}
\multicolumn{2}{c}{}& HA & HLH & HLP & B & BA & BLH & BLP & M & MA & MLH\\
\cmidrule(l){2-12}
\multirow{5}{*}{\rotatebox[origin=c]{90}{UDA method\hspace{0.8cm}}}
& DAN &  -14.29 & 11.59 & 27.03 & 5.26 & 5.26 & 6.85 & 7.79 & 8.82 & -13.33 & 1.47 \\ [0.2ex]
& JAN &  7.14 & 18.84 & 51.35 & 26.32 & 78.95 & 16.44 & 6.49 & 29.41 &  -6.67 & \textbf{7.35} \\ [0.2ex]
& DANN &  21.43 & 17.39 & 48.65 & 29.82 & 68.42 & 16.44 & 9.09 & 29.41 &  \textbf{20.00} & 5.88 \\ [0.2ex]
& ADDA &  \textbf{28.57} & 17.39 & 48.65 & 29.82 & \textbf{89.47} & \textbf{19.18} & 7.79 & 29.41 & 13.33 & \textbf{7.35} \\ [0.2ex]
& CDAN &  0 & 18.84 & 45.95 & 28.07 & 73.68 & 16.44 & 9.09 & \textbf{38.24} & -6.67 & \textbf{7.35} \\ [0.2ex]
& BSP & 14.29 & 18.84 & \textbf{75.68} &\textbf{ 31.58 }& 78.95 & 17.81 & 7.79 & 35.29 & 13.33 & \textbf{7.35} \\ [0.2ex]
& AFN & -21.43 & \textbf{20.29} & 54.05 & 28.07 & 84.21 & 15.07 & \textbf{11.69} & 26.47 &  6.67 & 4.41 \\ [0.2ex]
& MCC & 7.14 & \textbf{20.29} & 54.05 & 21.05 & 57.89 & 13.70 & 5.19 & 20.59 &  -6.67 & 4.41 \\ [0.2ex]
\cmidrule[0.9pt](l){2-12}
\end{tabular}
\caption{Comparison of change in AUPRC results for each UDA method with respect to the unadapted baseline method (\textit{Src}) for different datasets. H is the source dataset and the target datasets are listed in columns. The rows represent the improvements (in \%) for each UDA-method.}
\label{table:prauc_imp}
\end{table*}


\end{document}